\pdfoutput=1

\documentclass[11pt]{article}

\usepackage{ACL2023}

\usepackage{times}
\usepackage{latexsym}

\usepackage[T1]{fontenc}

\usepackage[utf8]{inputenc}

\usepackage{microtype}

\usepackage{inconsolata}
\usepackage{kotex}
\usepackage{multirow}
\usepackage{amsmath}
\usepackage{graphicx}
\usepackage{caption}
\usepackage{makecell}
\usepackage{hhline}
\usepackage{url}
\usepackage{amssymb}
\usepackage{algorithm}
\usepackage{algorithmic}
\usepackage{xcolor}
\usepackage[normalem]{ulem}
\usepackage{color}
\usepackage{vcell}
\usepackage{natbib}
\usepackage{diagbox}
\usepackage{booktabs}

\title{DSTEA: Improving Dialogue State Tracking \\ via Entity Adaptive Pre-training}

\author{Yukyung Lee$^1$, Takyoung Kim$^1$, Hoonsang Yoon$^1$, Pilsung Kang$^1$, Junseong Bang$^2$, Misuk Kim$^3$ \\
    $^1$Korea University, Seoul, Republic of Korea \\ 
    $^2$Electronics and Telecommunications Research Institute, Daejeon, Republic of Korea \\
    $^3$Sejong University, Seoul, Republic of Korea \\ 
    \texttt{\small $^1$yukyung\_lee, takyoung\_kim, hoonsang\_yoon, pilsung\_kang@korea.ac.kr}\\ 
    \texttt{\small $^2$hjbang21pp@gmail.com}\\
    \texttt{\small $^3$misuk.kim@sejong.ac.kr}}

\begin{document}
\maketitle
\begin{abstract}
Dialogue State Tracking (DST) is critical for comprehensively interpreting user and system utterances, thereby forming the cornerstone of efficient dialogue systems. Despite past research efforts focused on enhancing DST performance through alterations to the model structure or integrating additional features like graph relations, they often require additional pre-training with external dialogue corpora. In this study, we propose DSTEA, improving Dialogue State Tracking via Entity Adaptive pre-training, which can enhance the encoder through by intensively training key entities in dialogue utterances. DSTEA identifies these pivotal entities from input dialogues utilizing four different methods: ontology information, named-entity recognition, the \textit{spaCy}, and the \textit{flair} library. Subsequently, it employs selective knowledge masking to train the model effectively. Remarkably, DSTEA only requires pre-training without the direct infusion of extra knowledge into the DST model. This approach resulted in substantial performance improvements of four robust DST models on MultiWOZ 2.0, 2.1, and 2.2, with joint goal accuracy witnessing an increase of up to 2.69\% (from 52.41\% to 55.10\%). Further validation of DSTEA's efficacy was provided through comparative experiments considering various entity types and different entity adaptive pre-training configurations such as masking strategy and masking rate.
\end{abstract}
\section{Introduction}
\label{sec:Introduction}

A task-oriented dialogue system (TOD), which aims to complete a certain task in a specific domain, such as restaurant reservation, can be modularized into four sub-tasks: natural language understanding, dialogue state tracking (DST), dialogue policy learning, and natural language generation \cite{zhang2020recent}. Among them, DST, which seeks to track the structured belief state, plays an important role because the final quality of the entire dialogue system significantly depends on the accurate tracking of such belief states. However, as the dialogue system mainly focuses on the current dialogue turn, generating the correct belief state in a multi-turn dialogue is a challenging task \cite{kim-etal-2020-efficient}. 

There are two main directions of recent studies endeavoring to improve DST performance by generating the correct belief states: modifying the model structure or conducting additional pre-training using in-domain dialogue corpora. In the former, TRADE \citep{wu-etal-2019-transferable} and SOM-DST \citep{kim-etal-2020-efficient} were used in an attempt to accurately reflect the accumulated belief states by jointly training the encoder-decoder structure using a pointer network \citep{see-etal-2017-get}. In addition, SST \citep{Chen_Lv_Wang_Zhu_Tan_Yu_2020}, GCDST \citep{wu-etal-2020-gcdst}, and CSFN-DST \citep{zhu-etal-2020-efficient} were employed to provide rich information to the encoder by using the schema graph as an extra feature. In the latter, TOD-BERT \citep{wu-etal-2020-tod} and DialoGLUE \citep{MehriDialoGLUE2020} performed additional pre-training based on masked language modeling on the encoder model (e.g., BERT \cite{devlin-etal-2019-bert}). In particular, DialoGLUE learned a representation suitable for the target domain by taking the same pre-training strategy using a large amount of external dialogue corpora before fine-tuning \cite{MehriDialoGLUE2020, gururangan-etal-2020-dont}. In addition, models such as SimpleTOD \citep{hosseini2020simple} and SOLOIST \citep{peng-etal-2021-soloist} performed additional pre-training based on language modeling on the decoder model (e.g., GPT \cite{radford2019language}) to capture TOD-related features. 

Although the aforementioned models have shown some positive effects on DST performance improvement, we believe that they can be further improved by considering the following points. First, language model pre-training can be enhanced by employing a customized strategy for the dialogue domain for the DST. Conversational models such as DialoGPT \citep{zhang-etal-2020-dialogpt}, TOD-BERT, and DialogBERT \citep{gu2021dialogbert} were trained to capture textual and semantic features of dialogue context and showed significant performance improvement in response generation and response selection \cite{gu2021dialogbert}. Therefore, DST performance improvement can also be achieved when training is based on a pre-trained language model specialized in the dialogue domain. Second, utilizing additional techniques to capture task-related knowledge effectively will also be helpful \cite{zhang-zhao-2021-structural, gururangan-etal-2020-dont}. 
DialoGLUE learned representations suitable for the target domain by performing additional pre-training using seven different datasets across four tasks: intent prediction, slot filling, semantic parsing, and DST. Similarly, PPTOD \citep{su-etal-2022-multi} used 11 datasets to train four tasks: natural language understanding, DST, dialogue policy learning, and natural language generation. Although this learning strategy is effective for performance improvement, computational cost issues arise because it requires a large amount of dialogue corpora. Therefore, finding an effective as well as efficient pre-training method that captures task-related knowledge from a DST dataset is necessary.

Note again that the main purpose of DST is to accurately extract the value assigned to a specific domain-slot pair from the dialogue utterance. Such an information extraction task can emphasize the semantics of important information by pre-training focusing on the entity feature of the input sequence. Accordingly, the performance of various natural language processing tasks can be improved \cite{yang2021survey}. Hence, we focused on the point that previous studies have rarely considered entities that provide rich information from dialogue utterances \citep{wang-etal-2020-pre} (i.e., an entity that appears in one sentence, such as identifying people). In light of this, the approach we propose uniquely leverages entity information from the target dataset, which can help reduce the computational cost without significantly impacting performance. Continuing on this path, we introduce DSTEA (improving \underline{\textbf{D}}ialogue \underline{\textbf{S}}tate \underline{\textbf{T}}racking via \underline{\textbf{E}}ntity \underline{\textbf{A}}daptive pre-training) in this paper, a methodology in which the entity representations in DST models are intensively trained after extracting important information from a given utterance. Our approach is applicable to any BERT-based belief tracker and can enhance the tracking ability. We verified the effectiveness of the DSTEA using four strong DST models for various versions of MultiWOZ \cite{budzianowski-etal-2018-multiwoz}. Experimental results showed that our training strategy had a positive effect on the performance improvement of the DST model and demonstrated the usefulness of entity-level information in multi-turn dialogue. Furthermore, additional analysis of slot-meta (domain-slot pair) information, and value, showed that the proposed DST model affords a lower error rate of predicted values than the existing models.

\begin{figure}[t!]
\centering
\includegraphics[width=\columnwidth]{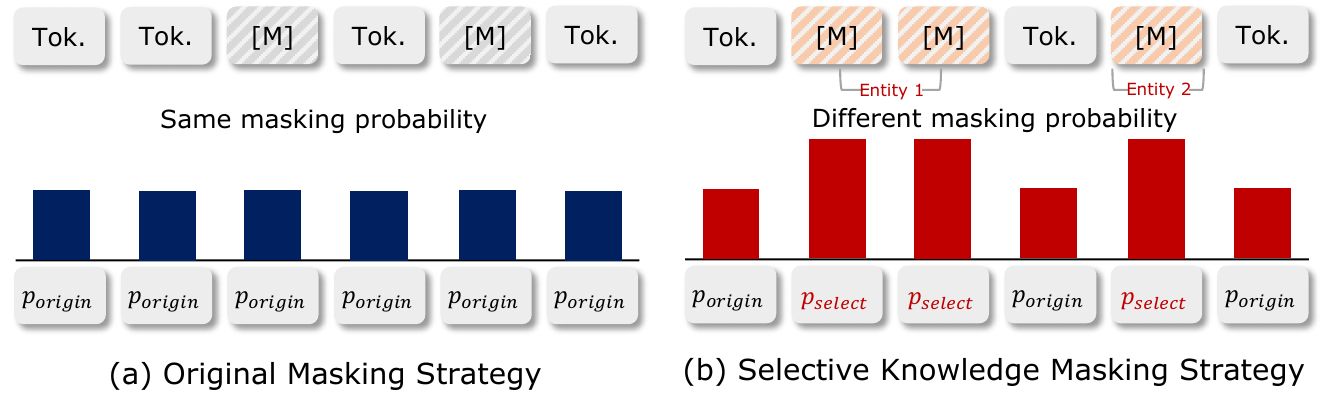}
\caption{Comparison between original and proposed masking strategies. We use selective knowledge for effective pre-training on the DST. We give the masking probability by distinguishing between knowledge selected as important information ($p_{select}$) or relatively insignificant information ($p_{origin}$).}
\label{fig:Masking_Strategy}
\end{figure}

\section{Related Work}
\label{sec:Related Work}

\begin{figure*}[t!]
\centering
\includegraphics[width=0.8\textwidth]{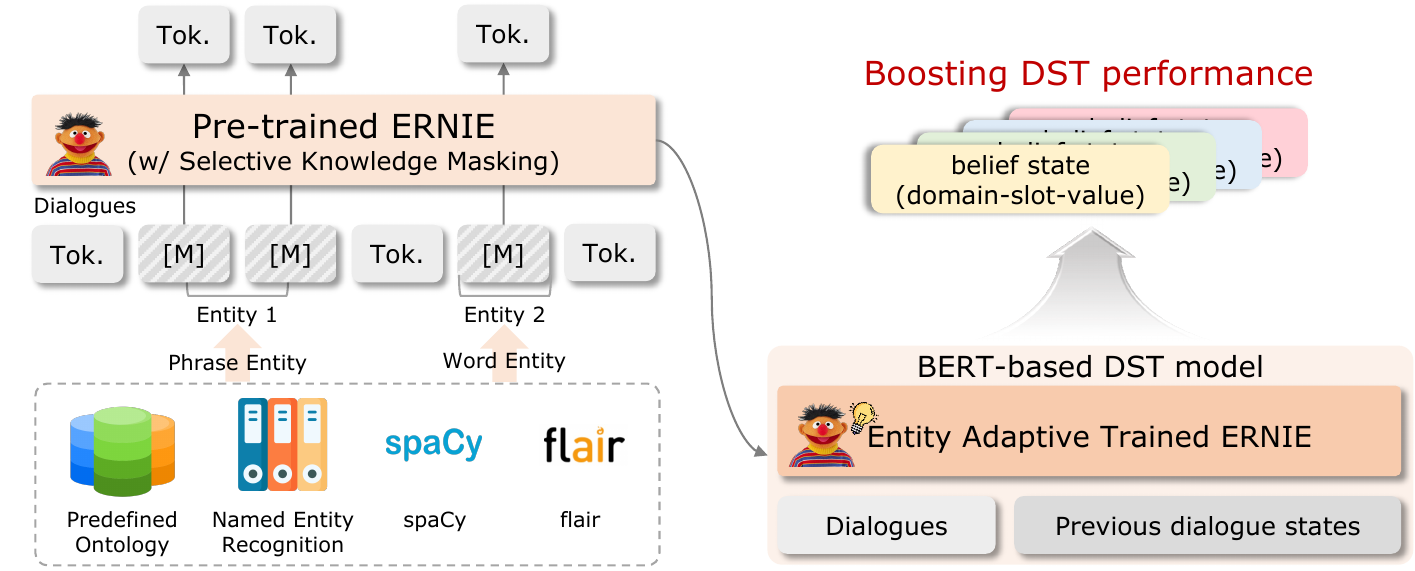}
\caption{The architecture of DSTEA. Our method enriches word and phrase entity information to encoder during further pre-training with selective knowledge masking.}
\label{fig:DSTEA_Overview}
\end{figure*}

\subsection{Task Adaptive Pre-training for DST}

Although large-scale pre-trained language models have achieved remarkable successes in various natural language processing tasks, models specialized for target domains and tasks have been continuously studied. Adaptive pre-training refers to a process in which a language model trained in the general domain is pre-trained to learn the knowledge suitable for a specific domain or task. \citet{gururangan-etal-2020-dont, beltagy-etal-2019-scibert, lee2020biobert} proposed a training strategy for a specific domain or task by adding an adaptation phase between pre-training and fine-tuning. In the area of DST, some studies have attempted to conduct task adaptive pre-training. DialoGLUE employed ConvBERT, which was trained with large amounts of open-domain dialogues and performed adaptive pre-training on the target dataset. In addition, ConvBERT-DG, which leveraged additional pre-training with seven DialoGLUE benchmarks, proved the surprising effect of self-supervised training. Further, SimpleTOD and SOLOIST performed pre-training through an auto-regressive objective with a GPT-based model and achieved high performance in DST. In particular, they achieved performance gain by training DST, action decision, and response generation together. In the case of \citet{zhao-etal-2021-effective-sequence},
Pegasus \citep{pmlr-v119-zhang20ae} pre-training objective were applied to T5 \citep{JMLR:v21:20-074}, and good performance was achieved without any pre/post-processing.

Inspired by the aforementioned methods, we propose a new adaptive pre-training method for DST. However, in contrast to existing methods, our strategy can enhance performance through a modified masking strategy to further pre-train the target dataset without any external dialogue corpora.

\subsection{Knowledge-enhanced Masking Strategy}

Knowledge can not only be decomposed into different levels of granularity but also be divided into unstructured and structured knowledge \cite{yang2021survey}. Unstructured knowledge comprises entities and text, whereas structured knowledge refers to a predefined structure such as a knowledge graph or syntax tree. This knowledge is transferred to the pre-trained model in the form of rich information and can improve the performance of downstream tasks \cite{yang2021survey, https://doi.org/10.48550/arxiv.2110.08455}. Pre-training methods such as BERT learn general-purpose knowledge through random uniform token masking \cite{levine2021pmimasking}, as illustrated in Figure\ref{fig:Masking_Strategy}-(a). However, this masking strategy trains information about a single segment \citep{joshi-etal-2020-spanbert} and has a limitation that models cannot learn related sub-word tokens together. Therefore, several studies modified the original masking strategy of BERT and attempted to improve performance in various natural language processing downstream tasks through appropriate knowledge injection \cite{yang2021survey}. ERNIE \citep{Sun_Wang_Li_Feng_Tian_Wu_Wang_2020} is a language model that contains entity information. It defines meaningful tokens, entities, and phrases as knowledge. Moreover, it was the first model to perform continual learning of high-level knowledge during the pre-training process. Further, SpanBERT \citep{joshi-etal-2020-spanbert} learns a continuous random span. Because the span boundary representation is learned without relying on individual tokens, it is more advanced than the original BERT. \citet{levine2021pmimasking} proposed pointwise mutual information masking that effectively conducts pre-training by simultaneously masking highly relevant n-gram tokens. \citet{Guu2020REALMRL} proposed REALM with salient-span masking to learn only named-entity and date information for question answering. Furthermore, \citet{roberts-etal-2020-much} considered entity information by applying salient-span masking to T5 \citep{JMLR:v21:20-074}; it showed high performance in a question answering task. However, the method of properly injecting entity-level knowledge for DST has been relatively less investigated.

In this study, we found that the knowledge definition of ERNIE and the entity masking of REALM are well suited to the nature of dialogues. Inspired by this finding, we propose selective knowledge masking to focus on the important entities. In summary, we attempted to capture information suitable for DST through a new entity masking strategy after extracting fine-grained knowledge entities from dialogues using an entity-specific ERNIE encoder.

\section{Proposed Method}

In this study, we propose DSTEA to intensively learn important knowledge about DST through entity adaptive pre-training. After extracting entities from the target dialogue data, DSTEA can capture DST-specific features by selectively training more focus on these entities.

The overall architecture of the DSTEA is shown in Figure \ref{fig:DSTEA_Overview}. In particular, we assume that the entity information appearing in the utterance is significant knowledge for DST and apply entity adaptive pre-training using a selective knowledge masking strategy, as shown in Figure \ref{fig:Masking_Strategy}-(b). After applying this pre-training, DSTEA learns belief state tracking by utilizing previous DST models, such as SOM-DST, Trippy \citep{heck2020trippy}, SAVN \citep{wang-etal-2020-slot} and STAR \citep{ye2021star}, which have achieved excellent performance and used BERT as an encoder.

\subsection{Entity Adaptive Pre-training}

\begin{figure}[t]
\centering
\includegraphics[width=0.8\columnwidth]{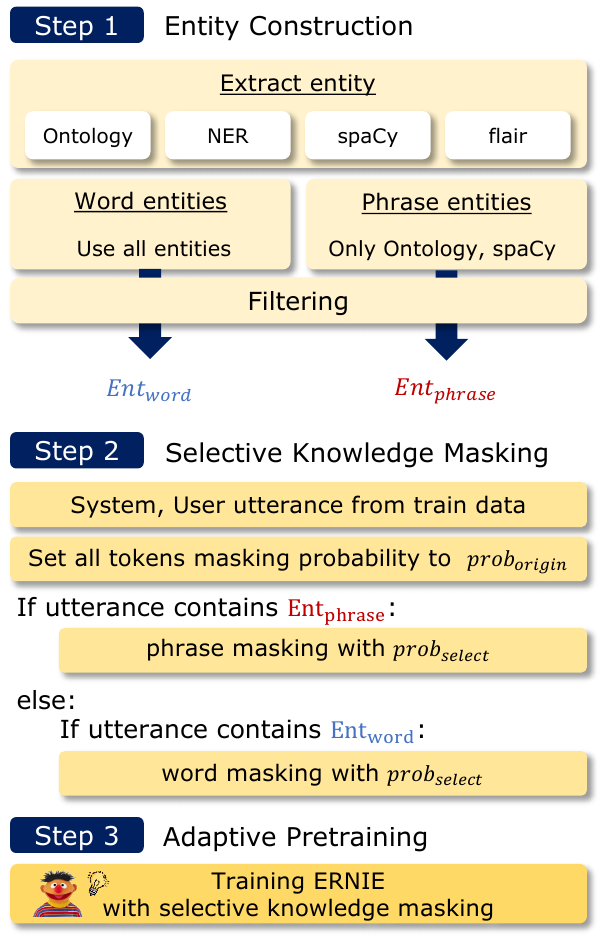}
\caption{Selective Knowledge Masking Procedure. The proposed masking strategy comprises three steps. In step 1, word and phrase entities are extracted and filtered through ontology, NER, \textit{spaCy}, and \textit{flair}. In step 2, selective knowledge masking is performed using the extracted entity set. In step 3, entity adaptive pre-training is performed based on the masking probability newly defined in step 2.}
\label{fig:selective knowledge masking_Procedure_1}
\end{figure}

The encoder architecture is an essential part of the DST model. The purpose of the encoder model in the proposed method is to learn an inductive bias suitable for DST during the pre-training process so that the representation of the pre-training model can be used to learn the dialogue information more accurately. The adaptive pre-training method proposed for DSTEA is shown in Figure \ref{fig:selective knowledge masking_Procedure_1}. Pre-training comprises three steps: entity set construction, selective knowledge masking, and adaptive pre-training. After extracting entities from the utterance, a higher masking rate is assigned to them,  while the original masking rate is assigned to the remaining tokens.

\subsubsection{Entity Construction}

One of the most important parts of entity adaptive pre-training is entity set construction. In this study, the entities were collected in four ways. First, entities were selected using ontology information. This is because an ontology is the most readily available form of information from the dataset and specifies the most important words. Second, after establishing a named-entity recognition (NER) model, the inference was conducted on the MultiWOZ dataset to collect the entities. An ERNIE-based entity tagging model was used, and the model is trained using the CONLL 2003 dataset \citep{tjong-kim-sang-de-meulder-2003-introduction}. Third, entities were extracted using the \textit{spaCy} \citep{spaCy2} library. The \textit{spaCy} entity recognizer extracts entities in span units, including entity types such as location, language, person, and product. Finally, entities were extracted using the \textit{flair} library \cite{akbik2019flair}. The \textit{flair} entity tagger is a model that is trained based on CONLL 2003 and extracts entities in span units. The entities extracted using these methods are a mixture of words and phrases. We attempted to learn these by distinguishing between word and phrase entities. In our methodology, the entities obtained through the four distinct strategies were consolidated into a unified set. This unified set served as a candidate pool for the masking tokens.

\textit{\textbf{Word Entity}} First, each extracted entity was split into word units to compose a word entity. To prevent overfitting during this process, information about time and numbers was excluded from the entity. Because random times and numbers are used when constructing dialogue datasets \citep{budzianowski-etal-2018-multiwoz}, the appearance of unseen information during a dialogue may prevent the DST model from responding correctly to unseen slots or values. In other words, if the entities are trained regarding time and numbers with selective knowledge masking, the biased model is highly likely to generate incorrect values. Therefore, these values were not considered. Additionally, stopwords from the \textit{NLTK} \citep{bird2009natural} library were used to exclude words that were not helpful for training. Moreover, filtering was performed when the punctuation mark was extracted as an entity.

\textit{\textbf{Phrase Entity}} To learn a phrase entity, cases that included an unknown token (i.e., [UNK]) in the phrase were excluded. Next, the phrase entities defined for each utterance were extracted in advance, and then pre-training was performed using randomly selected phrase entities. Even if a phrase entity included stopwords, filtering was not performed for masking in span units, and information about numbers and times was excluded, as for word entities. However, phrase entities extracted by the NER model and \textit{flair} were of poor quality; therefore, entities were extracted using only the ontology and \textit{spaCy}.

\subsubsection{Selective Knowledge Masking}

Selective knowledge masking is a method for learning important knowledge after selecting essential information from user and system utterances to inject an inductive bias suitable for DST. As shown in Figure \ref{fig:Masking_Strategy}, the previous DST model used an encoder such as BERT, which was trained by random masking without considering the characteristics of tokens. By contrast, we identified entities appearing in dialogues and assigned a higher masking rate to them while giving the original masking rate to non-entity tokens.

Selective knowledge masking is described in step 2 in Figure \ref{fig:selective knowledge masking_Procedure_1}. The one-turn utterance token sequence is $U_t\ =\ (tok_1,tok_2,\ldots,tok_L)$, while the defined word entity is $Ent_{word}\ =\ (w_1,w_2,\ldots,w_N)$, the phrase entity is $Ent_{phrase}\ =\ (P_1,P_2,\ldots,P_M)$, the total length of the token sequence is $L$, the total number of word entities is $N$, and the total number of phrase entities is $M$. All tokens appearing in $U_t$ have a masking probability, and whether each token is masked or not is determined by its masking probability before training a masked language model. $prob_{origin}$ refers to the masking probability for general tokens, whereas $prob_{select}$ refers to the masking probability for entity tokens. In the proposed DSTEA, $prob_{select}$ is always greater than $prob_{origin}$. The selective knowledge masking proposed in this study proceeds as follows. First, the masking probability of all tokens in every dialogue utterance is initialized to $prob_{origin}$. When a specific utterance includes a predefined entity, the masking probability of the entity token ($Ent_{word}$) or entity span ($Ent_{phrase}$) is changed to $prob_{select}$. After changing the masking probability of the token to $prob_{select}$, masking is performed in units of words and phrases.

\section{Experimental Settings}

\subsection{Datasets} 

Experiments were conducted using MultiWOZ 2.0, 2.1, and 2.2 \citep{budzianowski-etal-2018-multiwoz, eric2019multiwoz, zang2020multiwoz}, the most widely used datasets for DST. Table \ref{tab:data statistics} shows statistics on MultiWOZ 2.0, 2.1 and 2.2. Similar to previous DST models, the experiments were conducted using five domains, namely, restaurant, train, hotel, taxi, and attraction. We followed the preprocessing procedures for each of the four baselines, most of which were provided by TRADE.

\begin{table}[h!]
\centering
\resizebox{0.8\columnwidth}{!}{%
\setlength{\tabcolsep}{15pt}
\renewcommand{\arraystretch}{1.3}
\begin{tabular}{c|c|c|c} 
\Xhline{3\arrayrulewidth}
\multirow{2}{*}{\textbf{Domain}} & \multicolumn{3}{c}{\textbf{MultiWOZ 2.0, 2.1, 2.2}}  \\ 
\cline{2-4}
                                 & \textbf{train} & \textbf{dev} & \textbf{test}        \\ 
\hline
\textbf{attraction}              & 2,717           & 401          & 395                  \\ 
\textbf{hotel}                   & 3,381           & 416          & 394                  \\ 
\textbf{restaurant}              & 3,813           & 438          & 437                  \\ 
\textbf{taxi}                    & 1,654           & 207          & 195                  \\ 
\textbf{train}                   & 3,103           & 484          & 494                  \\
\Xhline{3\arrayrulewidth}
\end{tabular}}
\caption{Data statistics on MultiWOZ 2.0, 2.1 and 2.2. }
\label{tab:data statistics}
\end{table}

\subsection{Baseline Models}

Four encoder structure-based baseline models were employed to evaluate the effectiveness of DSTEA: SOM-DST \citep{kim-etal-2020-efficient}, Trippy \citep{heck2020trippy}, SAVN \citep{wang-etal-2020-slot}, and STAR \citep{ye2021star}. Each model was trained according to publicly released implementations in the standard train/dev/test split of MultiWOZ.

\textbf{SOM-DST} introduced a state operation prediction that maintains the value of the previous slot instead of newly generating the value of every slot in each dialogue turn. In this model, the size of dialogue states is fixed size and some of dialogue states are selectively overwritten.

\textbf{Trippy} used three types of copy modules and classification gates, enabling the model to find values in the context of a conversation or the predictions of the previous turn. 

\textbf{SAVN} utilized slot attention and value normalization. Slot attention improves span prediction performance by sharing information between slot and utterance, while value normalization can correct the extracted span based on ontology.

\textbf{STAR} utilized slot token and slot self-attention to capture a strong correlation between slots. These two self-attention operations learn the relationship between slots and values and find the value through distance-based slot value matching.

In this paper, experimental results for the SAVN and STAR models on the MultiWOZ 2.2 dataset are absent in this paper due to the incompatibility of these ontology-based models with the dataset structure, which lacks a predefined ontology file. The hyperparameters are described in detail in the Appendix.

\subsection{Adaptive Pre-training Settings}

We trained the `pre-trained ERNIE-2.0' on dialogue and used the huggingface transformers \citep{wolf-etal-2020-transformers}, with \texttt{`nghuyong/ernie-2.0-en'} as the ERNIE model. During the experiment, the masking probability $prob_{origin}$ was set to 0.2, and $prob_{select}$ was set to 0.4. 

However, Trippy uses the label map for post-process to modify the model's prediction values. An example of a label map is shown in Table \ref{tabl:label_map}. The label map is performed only in Trippy and plays a key role in improving prediction accuracy. However, the official trippy code only provides the label map for MultiWOZ2.1. Therefore, we conducted experiments with the same label map for MultiWOZ 2.0 and 2.2.

In addition, SAVN consists of Slot Attention and Value Normalization modules. In the original paper, the performance is measured according to the number of layers of the Value Normalization module; we used ${VN}^{1}$.

\begin{table}
\centering
\resizebox{\columnwidth}{!}{%
\setlength{\tabcolsep}{10pt}
\renewcommand{\arraystretch}{1.3}
\begin{tabular}{l|l} 
\Xhline{3\arrayrulewidth}
\textbf{Value label} & \textbf{Value label variants}                 \\ 
\hline
moderate             & moderately, reasonable, reasonably, ...      \\
expensive            & high end, high class, fancy, ...              \\
sports               & multiple sport, multi sports, ...             \\
kings lynn           & king's lynn, king's lynn train station, ...  \\
1                    & one, just me, for me, myself, alone, ...      \\
\Xhline{3\arrayrulewidth}
\end{tabular}}
\caption{Examples of label map.}
\label{tabl:label_map}
\end{table}

\subsection{Evaluation Metrics}
The performances of the baseline and the proposed models were evaluated according to  the three following metrics: joint goal accuracy, slot accuracy, and relative slot accuracy. Joint goal accuracy \citep{williams-etal-2013-dialog} is an evaluation metric that verifies whether the predicted belief state exactly matches the gold label. Slot accuracy \cite{williams-etal-2013-dialog} is an evaluation metric that identifies the accuracy of slots among the predicted dialogue states. Relative slot accuracy \cite{kim-etal-2022-mismatch} is a recently proposed metric that complements joint goal accuracy and slot accuracy. In contrast to slot accuracy, relative slot accuracy only focuses on the gold reference and predicted slots of the current dialogue instead of all predefined slots in slot accuracy.

\section{Experimental Results}

Since there are some differences in evaluating the baseline models in their source codes, we unified the evaluation processes using the same criteria to secure a fair comparison. 

\begin{table*}[t!]
\centering
\resizebox{\textwidth}{!}{%
\setlength{\tabcolsep}{10pt}
\renewcommand{\arraystretch}{1.5}
\begin{tabular}{l|ccc|ccc|ccc} 
\Xhline{3\arrayrulewidth}
\multirow{2}{*}{\textbf{Model}}  & \multicolumn{3}{c|}{\textbf{MultiWOZ 2.0}}                            & \multicolumn{3}{c|}{\textbf{MultiWOZ 2.1}}                            & \multicolumn{3}{c}{\textbf{MultiWOZ 2.2}}                              \\ 
\cline{2-10}
                        & \vcell{JGA}            & \vcell{SA}       & \vcell{RSA}      & \vcell{JGA}            & \vcell{SA}       & \vcell{RSA}      & \vcell{JGA}            & \vcell{SA}       & \vcell{RSA}       \\[-\rowheight]
                        & \printcellbottom       & \printcellbottom & \printcellbottom & \printcellbottom       & \printcellbottom & \printcellbottom & \printcellbottom       & \printcellbottom & \printcellbottom  \\ 
\hline
SOM-DST \cite{kim-etal-2020-efficient}                 & 51.60                  & 97.20            & 86.59            & 52.41                  & 97.34            & 86.94            & 53.71                  & 97.38            & 87.31             \\
\textbf{SOM-DST + DSTEA (ours)}  & \textbf{54.11}         & \textbf{97.40}   & \textbf{87.51}   & \textbf{\uline{55.10}} & \textbf{97.46}   & \textbf{87.79}   & \textbf{\uline{55.23}} & \textbf{97.42}   & \textbf{87.55}    \\ 
\hline
Trippy \cite{heck2020trippy}                 & 52.63                  & 97.13            & 86.56            & 52.63                  & 97.20            & 87.98            & 53.38                  & 97.20            & 88.45             \\
\textbf{Trippy + DSTEA (ours)}   & \textbf{52.96}         & \textbf{97.18}   & \textbf{86.80}   & \textbf{54.87}         & \textbf{97.33}   & \textbf{88.50}   &\textbf{54.05}                        &\textbf{97.31}                  &  \textbf{88.81}                 \\ 
\hline
SAVN \cite{wang-etal-2020-slot}                   & 53.90                  & \textbf{97.43}   & \textbf{86.33}   & 53.65                  & 95.47            & 87.61            & \multicolumn{3}{c}{\multirow{2}{*}{-}}                        \\
\textbf{SAVN + DSTEA (ours)}     & \textbf{54.19}         & \textbf{97.43}   & 86.32            & \textbf{54.75}         & \textbf{97.52}   & \textbf{87.94}   & \multicolumn{3}{c}{}                                          \\ 
\hline
STAR \cite{ye2021star}               & 54.75                  & 97.44            & 86.19            & 54.22                  & 97.48            & 87.49            & \multicolumn{3}{c}{\multirow{2}{*}{-}}                        \\
\textbf{STAR + DSTEA (ours)} & \textbf{\uline{55.53}} & \textbf{97.49}   & \textbf{86.43}   & \textbf{55.02}         & \textbf{97.57}   & \textbf{87.88}   & \multicolumn{3}{c}{}                                          \\
\Xhline{3\arrayrulewidth}
\end{tabular}}

\caption{Comparison of the proposed (+DSTEA) with four baseline models on the test sets of MultiWOZ 2.0, 2.1, and 2.2. The scores of joint goal accuracy (JGA), slot accuracy (SA), and relative slot accuracy (RSA) are computed. The underline indicates the best JGA for each dataset.}
\label{tab:Main Result}
\end{table*}

\subsection{DST Performance}

The performances of DSTEA with four baseline models using MultiWOZ 2.0, 2.1, and 2.2 datasets are presented in Table \ref{tab:Main Result}. The joint goal accuracy improved in all datasets for all baseline models. More specifically, STAR + DSTEA in MultiWOZ 2.0 and SOM-DST + DSTEA in MultiWOZ 2.1 and 2.2 recorded the best performance. In particular, when DSTEA's entity adaptive pre-training was applied to SOM-DST, the most remarkable performance improvement (+2.69) was recorded in MultiWOZ 2.0. These performance improvements confirm that an effective representation can be learned by training entities, essential information for the dialogue domain, more intensively than other tokens. Because joint goal accuracy accepts the prediction as correct only when the accurate belief state is predicted overall dialogue turns, it is the strictest evaluation metric. Significant performance improvement in terms of joint goal accuracy implies that the entity adaptive pre-training is sufficiently effective. In addition to joint goal accuracy, both slot accuracy and relative slot accuracy are also improved in all cases except for SAVN in MultiWOZ 2.0. 
To validate the performance enhancement achieved by DSTEA, a paired t-test was conducted as an empirical evaluation. The p-value for this test is 0.0014 for the JGA metric, which is significantly below the standard significance threshold of 0.05. This outcome validates that the observed performance enhancements following the implementation of DSTEA are statistically significant and reliable. As such, our model presents a robust and statistically validated improvement over previous methodologies.


Table \ref{tab:Domain Result} illustrates the domain-level performance of DSTEA in conjunction with various baseline models. By integrating DSTEA, we observed substantial improvements across the majority of domains. A detailed depiction of the results for all slots for each model can be found in \ref{sec:slot_accuracy_analysis}. However, it is crucial to note that the DSTEA application does not always lead to performance enhancements. In rare instances, we have observed a marginal decline in performance. This is particularly pronounced in the taxi domain of the Trippy model, where the integration of DSTEA results in a notable decrease in performance concerning the `taxi departure' and `taxi destination' slots. Our analysis of this decline led us to conclude that these two slot values were often inaccurately predicted in reverse. While DSTEA has considerably advanced the model's ability to recognize entity (value), we identified instances where the model confounded the correlation between slots and values. This finding highlights the importance of models having a comprehensive understanding of the broader context to capture the relationships between slots and values accurately. Hence, to achieve optimal model performance, it is imperative to maintain a balance between enhanced entity recognition and a precise understanding of slot-value relations. We anticipate that the insights derived from this observation will significantly contribute to the refinement of our methodology, resulting in further improvements in the performance of our model.

\begin{table}[t!]
\centering
\resizebox{\columnwidth}{!}{%
\setlength{\tabcolsep}{6pt}
\renewcommand{\arraystretch}{1.5}
\begin{tabular}{l|c|c|c|c|c} 
\Xhline{3\arrayrulewidth}
\multicolumn{1}{c|}{\multirow{2}{*}{\textbf{Model}}} & \multicolumn{5}{c}{\textbf{Domain}}                                                                            \\ 
\cline{2-6}
\multicolumn{1}{c|}{}                                & \textbf{Attraction} & \textbf{~ ~Hotel~ ~} & \textbf{Restaurant} & \textbf{~ ~Taxi~ ~} & \textbf{~ ~Train~ ~}  \\ 
\hline 
SOM-DST                                              & 68.19               & \textbf{49.22}       & 65.89               & 57.01               & 71.61                 \\
\multicolumn{1}{r|}{\textbf{~ ~+ DSTEA}}             & \textbf{70.65}      & 48.61                & \textbf{69.88}      & \textbf{58.57}      & \textbf{73.38}        \\ 
\hline
Trippy                                               & 72.17               & 43.45                & 68.48               & \textbf{39.21}      & 70.48                 \\
\multicolumn{1}{r|}{\textbf{+ DSTEA}}                & \textbf{73.92}      & \textbf{48.81}       & \textbf{69.68}      & 35.53               & \textbf{71.10}        \\ 
\hline
SAVN                                                 & 66.82               & 47.28                & 66.46               & 62.96               & \textbf{75.51}        \\
\multicolumn{1}{r|}{\textbf{+ DSTEA}}                & \textbf{66.87}      & \textbf{49.25}       & \textbf{69.28}      & \textbf{65.91}      & 74.46                 \\ 
\hline
STAR                                                 & \textbf{68.57}      & 49.00                & 67.28               & 63.28               & 72.50                 \\
\multicolumn{1}{r|}{\textbf{+ DSTEA}}                & 67.83               & \textbf{49.09}       & \textbf{68.42}      & \textbf{71.03}      & \textbf{74.32}        \\
\Xhline{3\arrayrulewidth}
\end{tabular}}
\caption{Domain-specific performance on the test sets of MultiWOZ 2.1. The score of joint goal accuracy for each domain is computed.}
\label{tab:Domain Result}
\end{table}

\subsection{Effectiveness of Adaptive Pre-training}

To verify the effectiveness of the proposed selective knowledge masking, its performance was compared by changing the masking probability of the original masked language modeling. This comparative experiment was conducted using the SOM-DST model, which showed the highest performance improvement in the abovementioned experimental results. Table \ref{tab:Adaptive pretraining Result} shows how the performance is affected by the adaptive pre-training setting. \textbf{\textsc{SOM-DST (ERNIE)}} indicates that the encoder of SOM-DST is changed to ERNIE, and \textbf{\textsc{+ Random Masking ($P_{origin}$ = $\alpha$)}} represents the case of using random masking probability $\alpha$ during adaptive pre-training. $P_{select}$ and $P_{origin}$ of \textbf{\textsc{Selective Knowledge Masking (ours)}} are the masking probabilities of the entity tokens and the masking probability of the remaining tokens, respectively. The results showed that our strategy yielded the best performance. Notably, the performance of SOM-DST was improved by simply applying ERNIE, implying that the ERNIE encoder itself assuredly helps to extract the correct belief state. With respect to the masking probability, the performance was lower than that of \textbf{\textsc{SOM-DST (ERNIE)}} when the masking probability was set to 0.2. An interesting observation is that the DST performance can be improved by only increasing the masking probability for all tokens. However, our selective masking strategy outperformed \textbf{\textsc{+ Random Masking ($P_{origin}$ = $0.4$)}}, validating the greater effectiveness of the proposed selective-knowledge-masking method compared with simply increasing the masking probability.


\begin{table}[t!]
\centering
\resizebox{\columnwidth}{!}{%
\setlength{\tabcolsep}{10pt}
\renewcommand{\arraystretch}{1.3}
\begin{tabular}{r|c|c} 
\Xhline{3\arrayrulewidth}
\multicolumn{1}{c|}{\multirow{2}{*}{\textbf{Model }}}                                                                 & \multicolumn{2}{c}{\textbf{MultiWOZ 2.1 }}  \\ 
\cline{2-3}
\multicolumn{1}{c|}{}                                                                                                 & \textbf{JGA}   & \textbf{SA}                \\ 
\hline
\multicolumn{1}{l|}{\textbf{SOM-DST (BERT)}}                                                                          & 52.41          & 97.34                      \\
\multicolumn{1}{l|}{\textbf{SOM-DST (ERNIE)}}                                                                         & 53.97          & 97.40                      \\ 
\hline
\begin{tabular}[c]{@{}r@{}}\textbf{+ Random Masking }\\\textbf{($P_{origin} =0.2$)}\end{tabular}                      & 52.70          & 97.27                      \\
\begin{tabular}[c]{@{}r@{}}\textbf{+ Random Masking }\\\textbf{($P_{origin} =0.4$)}\end{tabular}                      & 54.24          & 97.42                      \\
\begin{tabular}[c]{@{}r@{}}\textbf{+ Selective Knowledge Masking (ours) }\\\textbf{($P_{select} =0.4$, $P_{origin}=0.2$)}\end{tabular} & \textbf{55.10} & \textbf{97.46}             \\
\Xhline{3\arrayrulewidth}
\end{tabular}}
\caption{Performance comparison table for adaptive pre-training of MultiWOZ 2.1. The strategies for the three masking strategies were tested based on SOM-DST(ERNIE).}
\label{tab:Adaptive pretraining Result}
\end{table}


\begin{figure}[t!]
\centering
\includegraphics[width=\columnwidth]{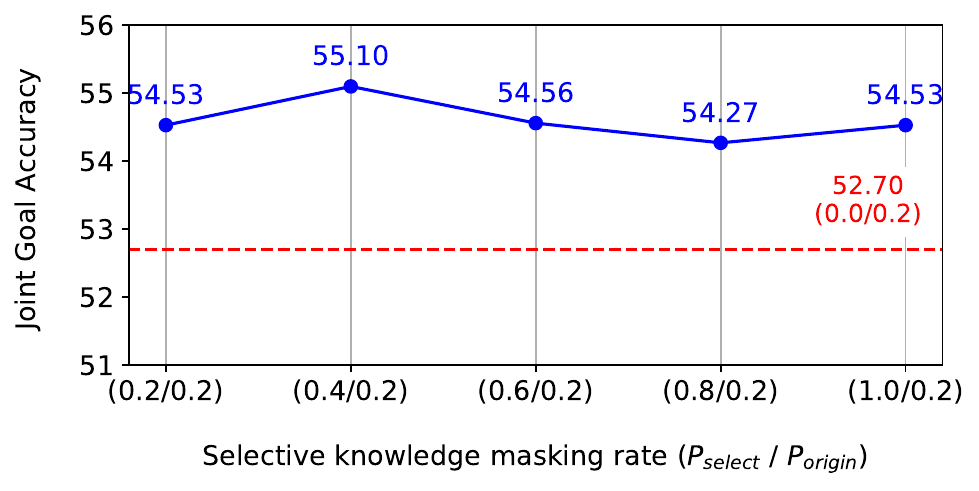}
\caption{Performance on DSTEA of MultiWOZ 2.1 test set with different selective knowledge masking rates.}
\label{fig:masking}
\end{figure}

\begin{table}
\centering
\resizebox{\columnwidth}{!}{%
\setlength{\tabcolsep}{5pt}
\renewcommand{\arraystretch}{1.3}
\begin{tabular}{c|c|c|c} 
\Xhline{3\arrayrulewidth}
\multirow{2}{*}{\textbf{Model }} & \multirow{2}{*}{\textbf{Type }}        & \multirow{2}{*}{\textbf{Entities }} & \textbf{MultiWOZ 2.1}  \\ 
\cline{4-4}
                                 &                                        &                                     & \textbf{JGA}          \\ 
\hline
\multirow{9}{*}{\textbf{\begin{tabular}[c]{@{}l@{}}SOM-DST\\~+ DSTEA\end{tabular}}} & \multirow{5}{*}{\textbf{Word Level }}  & \textbf{Ontology Only}              & 54.54                 \\
                                         &                                        & \textbf{Entity (spaCy) Only}        & 54.32                 \\
                                         &                                        & \textbf{Entity (NER) Only}          & 54.42                 \\
                                         &                                        & \textbf{Entity (flair) Only}        & 53.90                 \\
                                         &                                        & \textbf{Combine Words}              & \uline{54.79}                 \\ 
\cline{2-4}
                                 & \multirow{3}{*}{\textbf{Phrase Level}} & \textbf{Ontology Only}              & 54.47                 \\
                                 &                                        & \textbf{Entity (spaCy) Only}        & 54.31                 \\
                                 &                                        & \textbf{Combine Phrases}             & \uline{54.83}               \\ 
\cline{2-4}
                                 & \textbf{Combine All}                   & \textbf{Combine All}                & \textbf{55.10}        \\
\Xhline{3\arrayrulewidth}
\end{tabular}}
\caption{Performance according to the entity type of MultiWOZ 2.1. Comparison of individual performances of word and phrase entities according to the entity module. The underline indicates the highest score for each word-level and phrase-level entity type.}
\label{tab:Entity Type Result}
\end{table}

Figure \ref{fig:masking} shows the performance of DSTEA according to the change in the selective knowledge masking rate. The red dashed line represents the pre-training model that does not consider entities, which can be understood as the lower bound performance. The blue line indicates the performance of DSTEA according to different $P_{select}$ ratios. Selective knowledge masking clearly enhanced the DST performance regardless of the $P_{select}$ ratio. The best joint goal accuracy was reported when $P_{select} = 0.4$, but the worst case still yielded a significantly improved joint goal accuracy compared to that without selective knowledge masking.

\subsection{Effectiveness of Entity Types} 

In this section, we discuss how the entity extraction method affects the final DST performance. The entities were extracted using four modules, and both word and phrase units were considered. Table \ref{tab:Entity Type Result} shows the joint goal accuracy on MultiWOZ 2.1 with consideration of different entity types. Among the four individual extraction modules, ontology-based entity extraction was found to be the best for both word level and phrase levels. For word-level entities, other extraction methods also showed good performances, but the \textit{flair} library-based entity extraction afforded a slightly lower joint goal accuracy compared with the other three extraction modules. Note that when entities extracted by all four modules are combined, the joint goal accuracy was even improved than the single best extraction module for both word-level and phrase level. Moreover, when both word and phrase entities were aggregated, the best joint goal accuracy of 55.10 was achieved. Based on these results, entity set construction is also very important for DST performance improvement in addition to an appropriate learning strategy during the pre-training.


\begin{figure*}[t!]
\centering
\includegraphics[width=0.8\textwidth]{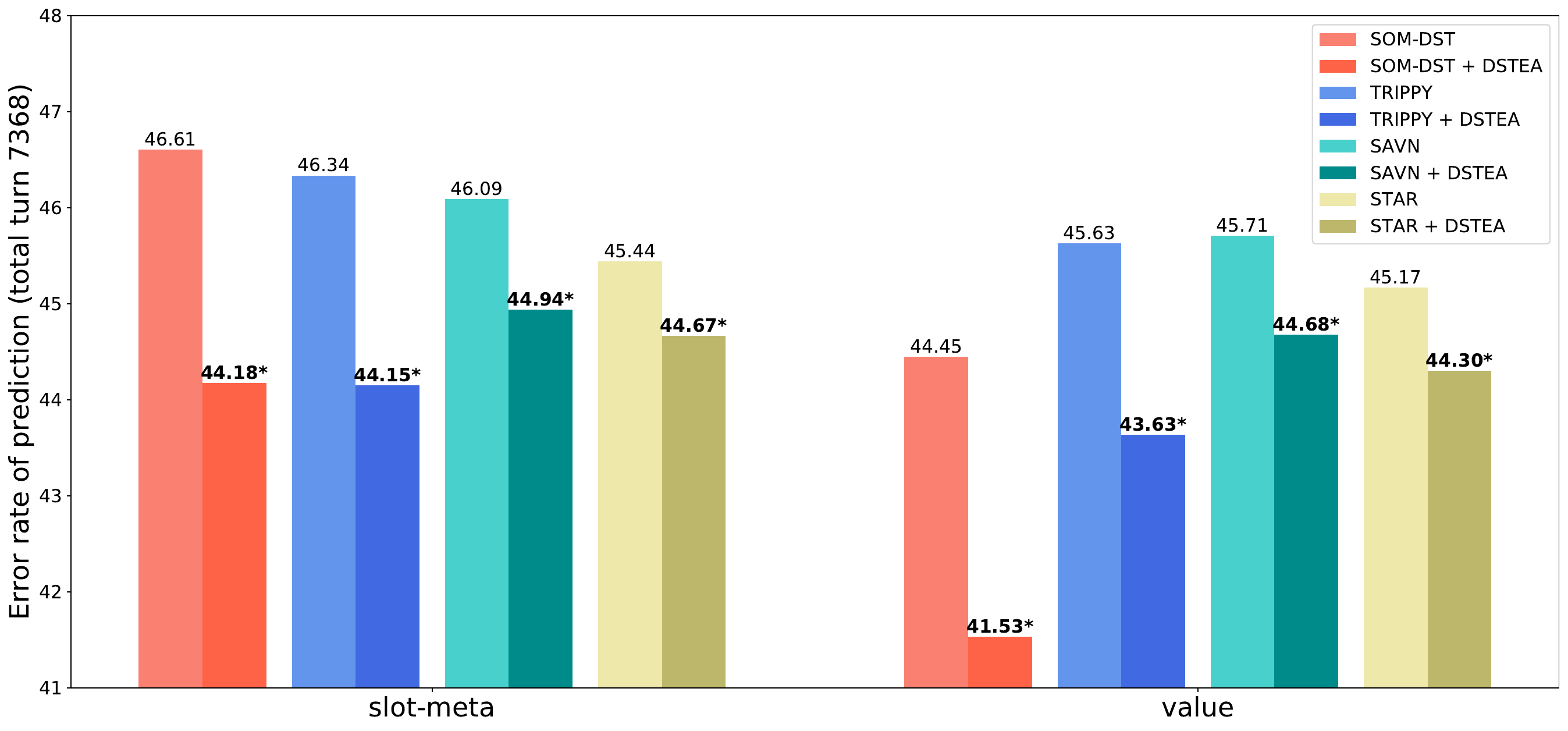}
\caption{Error rate between the baseline models (SOM-DST, Trippy, SAVN, STAR) and the DSTEA-applied models for slot-meta and value in MultiWOZ 2.1 (Total turns = 7368). The error rate is measured when there is a mismatch between the ground truth and the predicted value; thus, a lower value means an improved model.}
\label{fig:ErrorRate}
\end{figure*}

\subsection{Slot-meta and Value Error Rate}

In our work, the DST model generates predictions consisting of domain-slot pairs for each conversational turn. To examine the accuracy of these predictions, we introduced an `error rate' metric, which gives a comparative analysis of the error rates between our model DSTEA and baseline models concerning the slot-meta (domain-slot pair) and value. The error rate provides a granular understanding of where and to what extent the DST model could be erring, which is a perspective that the sole use of an accuracy metric might overlook. This detailed insight enables us to pinpoint and focus on specific areas that require improvement for enhancing the performance of model. As illustrated in Figure \ref{fig:ErrorRate}, DSTEA effectively reduces errors compared to the baseline models, attesting to its ability to accurately detect the domain and slot information and predict corresponding values. These findings suggest that the training methodology of DSTEA equips the model with an appropriate inductive bias, thereby augmenting the overall performance of the model.


\begin{figure}[t!]
\centering
\includegraphics[width=\columnwidth]{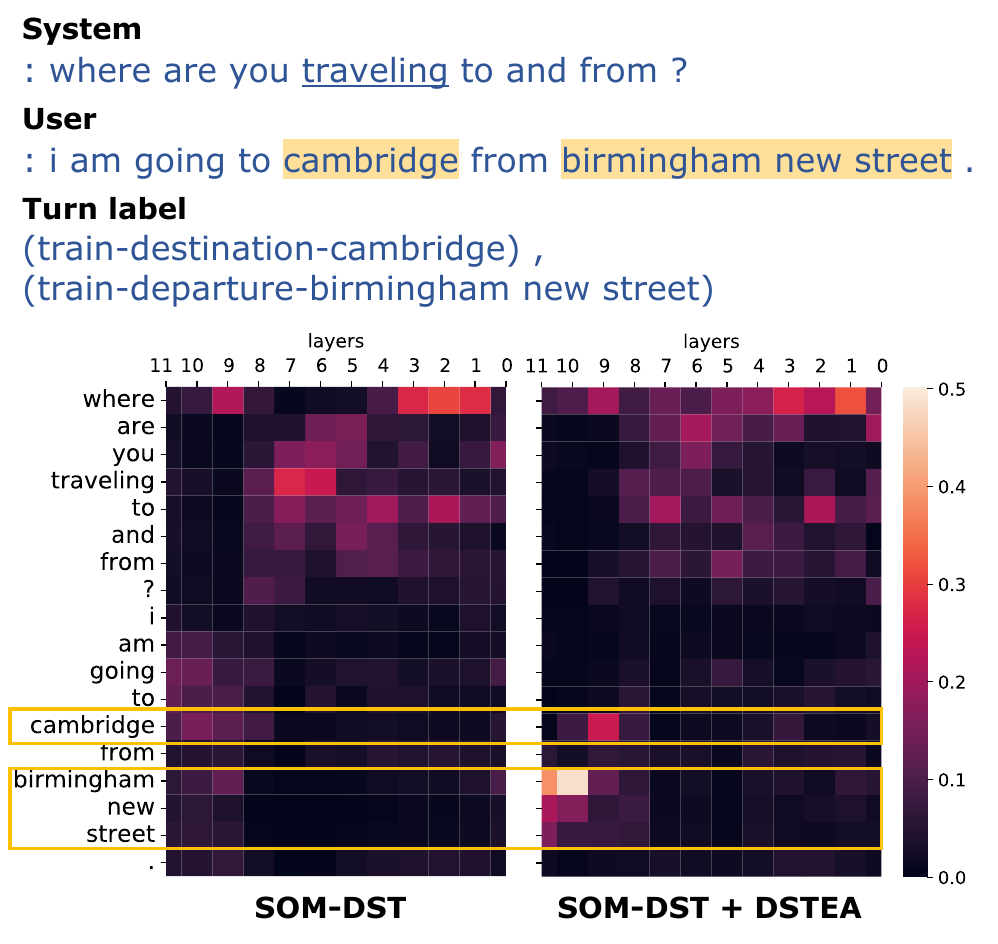}
\caption{Visualization of attention weights between SOM-DST and SOM-DST + DSTEA. The heatmap shows the average attention weight for all layers for the word 'traveling' (MultiWOZ 2.1 MUL0671 turn 1).}
\label{fig:Attention Score}
\end{figure}

\subsection{Attention Visualization}
In this part, we investigate the difference in attention weights between the baseline model and DSTEA. Figure \ref{fig:Attention Score} is a heatmap of the average attention weights for the word `traveling' when the first dialogue turn of MUL0671 is used as the input to both models. We can observe that much higher attention weights are assigned to the entities in the user utterance, especially to the proper nouns (`cambridge' and `birmingham'), by the DSTEA, supporting that the proposed entity adaptive pre-training results in focused concentration on informative entities. Note also that these higher attention weights appear in the upper layers of the encoder so that this important information is certainly delivered to the decoder in the case of SOM-DST to answer the slot-meta and value correctly.

\section{Conclusion}
In this study, we propose an entity adaptive pre-training framework,  named DSTEA, assuming that the essential knowledge of DST will be well captured if the pre-training of the language model focuses on informative entity tokens more intensively than others. In DSTEA, an entity-specialized language model, ERNIE, was employed for pre-training, while selective knowledge masking strategy is proposed to learn word and phrase entities more frequently than non-entities. Experimental results on four DST models show that the proposed DSTEA framework improved the baseline models in terms of JGA, SA, and RSA for three versions of the MultiWOZ dataset. Most entity extraction methods help to improve the DST performance, and combining the four extraction methods for both word and phrase entities yielded the best performance. We also verified that selective knowledge masking is more appropriate than simply increasing the masking rate to all types of tokens. The effect of DSTEA was also confirmed by the attention heatmap in that the informative entities were given higher attention weights with the DSTEA than without the DSTEA. We expect this pre-training method to be used effectively for DST under insufficient input data and entity-related tasks.

\bibliography{reference}
\bibliographystyle{acl_natbib}

\clearpage

\appendix

\section{Entity Extraction  Details}
\label{sec:entity details}
We use \textit{spaCy}, and \textit{flair} libraries to extract entities. The settings we used are below.

\begin{table}[h!]
\centering
\resizebox{0.6\columnwidth}{!}{%
\setlength{\tabcolsep}{21pt}
\renewcommand{\arraystretch}{1.2}
\begin{tabular}{c|c} 
\Xhline{3\arrayrulewidth}
\textbf{Setting} & \textbf{Assignment}  \\ 
\hline
model            & `en-core-web-sm'     \\ 
\hline
version          & 3.0.2                \\
\Xhline{3\arrayrulewidth}
\end{tabular}}
\caption{Settings for spaCy entity extraction.}
\end{table}

\begin{table}[h!]
\centering
\resizebox{0.6\columnwidth}{!}{%
\setlength{\tabcolsep}{15pt}
\renewcommand{\arraystretch}{1.2}
\begin{tabular}{c|c} 
\Xhline{3\arrayrulewidth}
\textbf{Setting} & \textbf{Assignment}  \\ 
\hline
model            & `flair/ner-english-large'    \\ 
\hline
version          & 0.10                \\
\Xhline{3\arrayrulewidth}
\end{tabular}}
\caption{Settings for flair entity extraction.}
\end{table}

\section{Evaluation}

To evaluate the effectiveness of DSTEA, we used joint goal accuracy (JGA), slot accuracy (SA), and relative slot accuracy (RSA). 
When the domain is $D$, the slot is $S$, and the value is $V$, the belief state $P$ predicted through the model can be defined as $P=(D, S, V)$. Also, when defining the goal state representing the correct answer as $G$, it can be expressed as $G=(D^{*}, S^{*}, V^{*})$. In addition, $T$ indicates the total number of slots. $M$ represents the number of missed slots that the model does not accurately predict among the slots included in the goal state $G$, and $W$ denotes the number of incorrectly predicted slots among the slots that do not exist in the gold state. $T^\ast$ denotes the number of unique slots that appear in prediction $P$ and goal status $G$ on a given turn.

Equation \ref{eq:jga}, \ref{eq:sa}, and \ref{eq:rsa} express JGA, SA, and RSA, respectively.
\begin{equation}
\label{eq:jga}
JGA = \left\{ 
  \begin{array}{ c l }
    1 & \, \textrm{if $P$}=\textrm{$G$}  \\
    0 & \, \textrm{otherwise}
  \end{array}
\right.
\end{equation}

\begin{equation}
\label{eq:sa}
    SA = {T-M-W \over T}
\end{equation}

\begin{equation}
\label{eq:rsa}
    RSA = {T^\ast - M -W \over T^\ast} \mbox{, where $0$ if $T^\ast=0$}
\end{equation}

In order to measure performance with the same standard, all predicted values obtained from each model were stored and evaluated using the same standard. Also, the result value is predicted every turn, but most slots have `none' value. Therefore, the performance was measured according to the convention of TRADE and SOM-DST, except when $V$ of $P$ and $V^{*}$ of $G$ were predicted to be `none' at the same time. If `none' is not excluded, all result values are predicted for all slots, so both SA and RSA are very close to 1 (ex 0.99), which is overestimated and may be incorrectly measured.

\section{Additional experiments with Knowledge Enhanced Encoders}

\begin{table}[t!]
\centering
\resizebox{0.8\columnwidth}{!}{%
\setlength{\tabcolsep}{13pt}
\renewcommand{\arraystretch}{1.3}
\begin{tabular}{r|c|c} 
\Xhline{3\arrayrulewidth}
\multicolumn{1}{c|}{\multirow{2}{*}{\textbf{Model }}}                                                                 & \multicolumn{2}{c}{\textbf{MultiWOZ 2.1 }}  \\ 
\cline{2-3}
\multicolumn{1}{c|}{}                                                                                                 & \textbf{JGA}   & \textbf{SA}                \\ 
\hline
\multicolumn{1}{l|}{\textbf{SOM-DST (BERT)}}                                                                          & 52.41          & 97.34                      \\
\multicolumn{1}{l|}{\textbf{SOM-DST (REALM)}}                                                                         & 51.82          & 97.19                      \\ 
\multicolumn{1}{l|}{\textbf{SOM-DST (ERNIE)}}                                                                         & 53.97          & 97.40                      \\ 
\hline
\multicolumn{1}{l|}{\textbf{SOM-DST + DSTEA}}                                                                         & \textbf{55.10}          & \textbf{97.46}                      \\ 
\Xhline{3\arrayrulewidth}
\end{tabular}}
\caption{Additional experiments with knowledge enhanced encoder (REALM, ERNIE) and DSTEA}
\label{fig: Knowledge Encoder}
\end{table}

As shown in Figure \ref{fig: Knowledge Encoder}, we conducted experiments to compare the performance of DSTEA with knowledge-enhanced encoders such as ERNIE and REALM, which were instrumental in inspiring our SKM method. Our results revealed that when REALM was employed as the backbone model for DST, it achieved relatively lower performance metrics, with a JGA of 51.82 and SA of 97.19. Notably, despite its documented efficacy in open-domain question-answering tasks, REALM did not manifest substantial improvements when fine-tuned for DST tasks. This discrepancy in performance, despite the use of a similar masking strategy, can be attributed to the primary design goal of REALM being centered around open-domain QA. The results suggest that while REALM is potent in its niche of open-domain QA tasks, it presents certain limitations when repurposed as a pre-training model for DST tasks. This experiment corroborates the effectiveness of our proposed DSTEA model, demonstrating its versatility and superior performance compared to the conventional knowledge-enhanced encoders in DST tasks.

\section{Training Details}
\label{sec:trainingdetails}

\begin{table*}[ht!]
\resizebox{\textwidth}{!}{%
\setlength{\tabcolsep}{3pt}
\renewcommand{\arraystretch}{1.5}
\centering
\begin{tabular}{l|c|c|c|cc|c|c} 
\Xhline{3\arrayrulewidth}
\multirow{2}{*}{\diagbox{\begin{tabular}[c]{@{}l@{}}\\Hyperparameters\end{tabular}}{Model}} & \multicolumn{2}{c|}{SOM-DST~\textbf{\textbf{\textbf{\textbf{+ DSTEA}}}}} & Trippy~\textbf{\textbf{\textbf{\textbf{+ DSTEA}}}} & \multicolumn{2}{c|}{SAVN~\textbf{\textbf{\textbf{\textbf{+ DSTEA}}}}}                                                                         & \multicolumn{2}{c}{STAR +\textbf{\textbf{\textbf{\textbf{~DSTEA}}}}}  \\ 
\cline{2-8}
                                                                                            & Encoder                  & Decoder                                       & Encoder                                            & \multicolumn{1}{c|}{\begin{tabular}[c]{@{}c@{}}Slot \\Attention\end{tabular}} & \begin{tabular}[c]{@{}c@{}}Value \\Normalization\end{tabular} & Encoder                  & Decoder                                    \\ 
\hline
Epochs                                                                                      & \multicolumn{2}{c|}{30}                                                  & 10                                                 & 5                                                                             & 3                                                             & \multicolumn{2}{c}{10 }                                               \\
Batch Size                                                                                  & \multicolumn{2}{c|}{64 }                                                 & 48                                                 & \multicolumn{2}{c|}{8 }                                                                                                                       & \multicolumn{2}{c}{16 }                                               \\
Learning Rate                                                                               & \multicolumn{1}{c}{4e-5} & 2e-4                                          & 1e-4                                               & 5e-5                                                                          & 1e-4                                                          & \multicolumn{1}{c}{3e-5} & 2e-4                                       \\
Drop Out                                                                                    & \multicolumn{2}{c|}{0.1 }                                                & 0.3                                                & \multicolumn{2}{c|}{0.1}                                                                                                                      & \multicolumn{2}{c}{0.1 }                                              \\
Warmup Ratio                                                                                & \multicolumn{2}{c|}{0.1}                                                 & 0.1                                                & \multicolumn{2}{c|}{0.1}                                                                                                                      & \multicolumn{2}{c}{0.1}                                               \\
Max Sequence Length                                                                         & \multicolumn{2}{c|}{256}                                                 & 180                                                & 32                                                                            & 512                                                           & \multicolumn{2}{c}{512}                                               \\
\Xhline{3\arrayrulewidth}
\end{tabular}}
\caption{Hyperparameters for DSTEA with baselines.}
\label{tab:hyperparameters}
\end{table*}

Our models are built with the huggingface transformers. And we trained our model with NVIDIA GeForce RTX 3090. Table \ref{tab:EntityAdaptivePretraining} shows hyperparameters used in entity adaptive pre-training and Table \ref{tab:hyperparameters} indicates DSTEA hyperparameters.

In addition, all baseline performances were conducted in the same environment as the DST learning. Baseline applied the same hyperparameter setting as the published code and the author. Table \ref{tab:hyperparameters} shows hyperparameters used in DSTEA with baselines.

\begin{table}[h!]
\centering
\resizebox{0.8\columnwidth}{!}{%
\setlength{\tabcolsep}{15pt}
\renewcommand{\arraystretch}{1.3}
\begin{tabular}{c|c} 
\Xhline{3\arrayrulewidth}
\textbf{Hyperparameter} & \textbf{Assignment}  \\ 
\hline
number of epoch         & 10                   \\ 
batch size              & 16                   \\ 
learning rate           & 6e-5                 \\ 
optimizer               & AdamW                \\ 
weight decay            & 0.01                 \\ 
warmup ratio            & 0.06                 \\ 
selective masking ratio & 0.4                  \\ 
original masking ratio  & 0.2                  \\
\Xhline{3\arrayrulewidth}
\end{tabular}}
\caption{Hyperparameters for entity adaptive pre-training.}
\label{tab:EntityAdaptivePretraining}
\end{table}

\section{Slot Accuracy Analysis}
\label{sec:slot_accuracy_analysis}

In this section, MultiWOZ 2.0, 2.1, and 2.2 reported all prediction accuracy results for each slot. Table \ref{tab:domain_2.0} is MultiWOZ 2.0, Table \ref{tab:domain_2.1} is MultiWOZ 2.1, Table \ref{tab:domain_2.2} is the joint goal accuracy performance for each slot for MultiWOZ 2.2. indicates. MultiWOZ 2.2 did not conduct experiments on SAVN and STAR implemented based on MultiWOZ 2.0 and 2.1 because the ontology was changed.

In these experimental results, it was confirmed that the all of 30 slots did not show high performance improvement, but most slot specific accuracy was similar or significantly improved. However, it was found that the performance decreased in `hotel-internet', which has values such as `yes' and `no' as the correct answer, and `restaurant-book people', which has numeric information as the correct answer. The following categorical slots are difficult to learn sufficiently through entity adaptive pre-training. It is expected that the methodology can be strengthened by proposing additional modules that can improve the performance of categorical slots in future work.

\begin{table*}
\centering
\resizebox{\textwidth}{!}{%
\setlength{\tabcolsep}{10pt}
\renewcommand{\arraystretch}{1.5}
\begin{tabular}{l|cc|cc|cc|cc} 
\Xhline{3\arrayrulewidth}
\multicolumn{9}{c}{MultiWOZ 2.0} \\ 
\hline
\vcell{Slot}           & \multicolumn{1}{l}{\vcell{SOM-DST}} & \multicolumn{1}{r|}{\vcell{\begin{tabular}[b]{@{}r@{}}SOM-DST\textbf{\textbf{\textbf{\textbf{}}}}\\\textbf{\textbf{\textbf{\textbf{+ DSTEA}}}}\end{tabular}}} & \multicolumn{1}{l}{\vcell{Trippy}} & \multicolumn{1}{l|}{\vcell{\begin{tabular}[b]{@{}l@{}}Trippy \\\textbf{\textbf{\textbf{\textbf{+ DSTEA}}}}\end{tabular}}} & \multicolumn{1}{l}{\vcell{SAVN}}  & \multicolumn{1}{l|}{\vcell{\begin{tabular}[b]{@{}l@{}}SAVN \\\textbf{\textbf{\textbf{\textbf{+ DSTEA}}}}\end{tabular}}} & \multicolumn{1}{l}{\vcell{STAR}}  & \multicolumn{1}{l}{\vcell{\begin{tabular}[b]{@{}l@{}}STAR\\\textbf{\textbf{\textbf{\textbf{+ DSTEA}}}}\end{tabular}}}  \\[-\rowheight]
\printcellbottom       & \multicolumn{1}{l}{\printcelltop}   & \multicolumn{1}{r|}{\printcellmiddle}                                                                                                                         & \multicolumn{1}{l}{\printcelltop}  & \multicolumn{1}{l|}{\printcellmiddle}                                                                                     & \multicolumn{1}{l}{\printcelltop} & \multicolumn{1}{l|}{\printcellmiddle}                                                                                   & \multicolumn{1}{l}{\printcelltop} & \multicolumn{1}{l}{\printcellmiddle}                                                                                   \\ 
\hline
attraction-area        & 92.70                               & 92.27                                                                                                                                                         & 91.66                              & \textbf{92.03}                                                                                                            & 88.52                             & \textbf{89.04}                                                                                                          & 89.80                             & \textbf{90.46}                                                                                                         \\
attraction-name        & 81.56                               & \textbf{85.34}                                                                                                                                                & 63.36                              & \textbf{65.64}                                                                                                            & 58.16                             & 57.44                                                                                                                   & 58.36                             & 57.40                                                                                                                  \\
attraction-type        & 89.33                               & \textbf{90.96}                                                                                                                                                & 82.78                              & \textbf{83.23}                                                                                                            & 81.88                             & \textbf{83.27}                                                                                                          & 87.22                             & 87.02                                                                                                                  \\
hotel-area             & 88.31                               & \textbf{89.42}                                                                                                                                                & 84.50                              & 82.70                                                                                                                     & 84.26                             & 82.42                                                                                                                   & 85.05                             & 81.19                                                                                                                  \\
hotel-book day         & 96.66                               & 96.27                                                                                                                                                         & 85.86                              & \textbf{87.96}                                                                                                            & 91.34                             & \textbf{91.87}                                                                                                          & 89.70                             & \textbf{92.32}                                                                                                         \\
hotel-book people      & 96.79                               & \textbf{97.57}                                                                                                                                                & 83.57                              & \textbf{84.93}                                                                                                            & 92.24                             & \textbf{92.43}                                                                                                          & 91.24                             & \textbf{94.18}                                                                                                         \\
hotel-book stay        & 97.03                               & \textbf{97.73}                                                                                                                                                & 84.52                              & \textbf{87.97}                                                                                                            & 92.02                             & \textbf{92.23}                                                                                                          & 90.71                             & \textbf{93.28}                                                                                                         \\
hotel-internet         & 85.06                               & \textbf{86.76}                                                                                                                                                & 81.96                              & 80.62                                                                                                                     & 84.26                             & 83.97                                                                                                                   & 83.87                             & 81.67                                                                                                                  \\
hotel-name             & 90.30                               & \textbf{91.73}                                                                                                                                                & 73.93                              & 72.90                                                                                                                     & 64.34                             & \textbf{65.39}                                                                                                          & 60.11                             & 57.91                                                                                                                  \\
hotel-parking          & 87.88                               & 86.76                                                                                                                                                         & 76.86                              & \textbf{78.80}                                                                                                            & 81.93                             & \textbf{84.63}                                                                                                          & 80.82                             & \textbf{83.72}                                                                                                         \\
hotel-pricerange       & 91.66                               & 90.11                                                                                                                                                         & 89.36                              & 87.49                                                                                                                     & 88.98                             & \textbf{89.31}                                                                                                          & 89.61                             & \textbf{90.10}                                                                                                         \\
hotel-stars            & 90.82                               & 90.31                                                                                                                                                         & 89.28                              & 88.70                                                                                                                     & 88.72                             & \textbf{88.84}                                                                                                          & 88.25                             & 87.88                                                                                                                  \\
hotel-type             & 79.65                               & \textbf{81.05}                                                                                                                                                & 64.34                              & \textbf{68.71}                                                                                                            & 72.92                             & 70.57                                                                                                                   & 69.01                             & \textbf{71.34}                                                                                                         \\
restaurant-area        & 94.49                               & 94.44                                                                                                                                                         & 88.73                              & \textbf{89.61}                                                                                                            & 89.23                             & \textbf{90.55}                                                                                                          & 89.95                             & 87.29                                                                                                                  \\
restaurant-book day    & 98.93                               & 98.09                                                                                                                                                         & 91.14                              & \textbf{91.99}                                                                                                            & 94.74                             & 94.62                                                                                                                   & 93.29                             & 92.97                                                                                                                  \\
restaurant-book people & 98.12                               & 97.97                                                                                                                                                         & 91.30                              & 90.31                                                                                                                     & 95.37                             & 95.24                                                                                                                   & 95.43                             & \textbf{95.50}                                                                                                         \\
restaurant-book time   & 97.48                               & 96.03                                                                                                                                                         & 90.50                              & \textbf{91.52}                                                                                                            & 93.69                             & 93.50                                                                                                                   & 90.63                             & 90.08                                                                                                                  \\
restaurant-food        & 94.27                               & \textbf{94.75}                                                                                                                                                & 93.47                              & \textbf{93.54}                                                                                                            & 93.27                             & 92.47                                                                                                                   & 93.15                             & 92.86                                                                                                                  \\
restaurant-name        & 86.13                               & \textbf{87.05}                                                                                                                                                & 69.87                              & 66.31                                                                                                                     & 52.01                             & 51.35                                                                                                                   & 52.99                             & \textbf{56.12}                                                                                                         \\
restaurant-pricerange  & 93.79                               & \textbf{93.89}                                                                                                                                                & 90.41                              & 89.26                                                                                                                     & 89.98                             & 89.92                                                                                                                   & 88.72                             & 88.38                                                                                                                  \\
taxi-arriveby          & 92.70                               & 86.69                                                                                                                                                         & 70.67                              & \textbf{72.58}                                                                                                            & 73.11                             & \textbf{73.55}                                                                                                          & 68.77                             & \textbf{73.22}                                                                                                         \\
taxi-departure         & 81.04                               & 80.34                                                                                                                                                         & 45.84                              & \textbf{48.45}                                                                                                            & 75.04                             & \textbf{75.86}                                                                                                          & 77.38                             & \textbf{77.81}                                                                                                         \\
taxi-destination       & 81.77                               & \textbf{85.17}                                                                                                                                                & 54.42                              & 53.93                                                                                                                     & 76.05                             & 74.88                                                                                                                   & 74.29                             & \textbf{80.20}                                                                                                         \\
taxi-leaveat           & 85.06                               & \textbf{88.71}                                                                                                                                                & 74.33                              & 72.56                                                                                                                     & 72.01                             & \textbf{74.20}                                                                                                          & 62.50                             & \textbf{74.10}                                                                                                         \\
train-arriveby         & 87.61                               & \textbf{87.92}                                                                                                                                                & 92.56                              & \textbf{94.14}                                                                                                            & 93.52                             & 93.37                                                                                                                   & 92.39                             & 89.42                                                                                                                  \\
train-bookpeople       & 89.65                               & 88.60                                                                                                                                                         & 80.62                              & \textbf{80.70}                                                                                                            & 88.47                             & 86.84                                                                                                                   & 79.66                             & \textbf{82.93}                                                                                                         \\
train-day              & 98.63                               & 98.56                                                                                                                                                         & 94.41                              & \textbf{95.31}                                                                                                            & 96.34                             & 96.12                                                                                                                   & 96.83                             & 96.64                                                                                                                  \\
train-departure        & 96.71                               & 96.71                                                                                                                                                         & 93.32                              & \textbf{93.66}                                                                                                            & 93.84                             & 93.14                                                                                                                   & 94.52                             & \textbf{95.08}                                                                                                         \\
train-destination      & 98.40                               & 97.73                                                                                                                                                         & 93.94                              & \textbf{94.98}                                                                                                            & 94.65                             & 93.89                                                                                                          & 95.16                             & \textbf{96.25}                                                                                                         \\
train-leaveat          & 69.29                               & \textbf{69.93}                                                                                                                                                         & 81.40                              & 80.47                                                                                                                     & 83.02                             & 82.69                                                                                                                   & 82.87                             & 76.64                                                                                                                  \\
\Xhline{3\arrayrulewidth}
\end{tabular}}
\caption{Slot-specific performance on the test sets of MultiWOZ 2.0. The score of joint goal accuracy for each slot is computed. If the \textbf{+ DSTEA} performance is improved compared to the baseline, it is shown in bold. }
\label{tab:domain_2.0}
\end{table*}

\begin{table*}
\centering
\resizebox{\textwidth}{!}{%
\setlength{\tabcolsep}{10pt}
\renewcommand{\arraystretch}{1.5}
\begin{tabular}{l|cc|cc|cc|cc} 
\Xhline{3\arrayrulewidth}
\multicolumn{9}{c}{\textbf{MultiWOZ 2.1}}\\ 
\hline
\vcell{Slot}           & \multicolumn{1}{l}{\vcell{SOM-DST}} & \multicolumn{1}{r|}{\vcell{\begin{tabular}[b]{@{}r@{}}SOM-DST\textbf{\textbf{\textbf{\textbf{}}}}\\\textbf{\textbf{\textbf{\textbf{+ DSTEA}}}}\end{tabular}}} & \multicolumn{1}{l}{\vcell{Trippy}} & \multicolumn{1}{l|}{\vcell{\begin{tabular}[b]{@{}l@{}}Trippy \\\textbf{\textbf{\textbf{\textbf{+ DSTEA}}}}\end{tabular}}} & \multicolumn{1}{l}{\vcell{SAVN}}  & \multicolumn{1}{l|}{\vcell{\begin{tabular}[b]{@{}l@{}}SAVN \\\textbf{\textbf{\textbf{\textbf{+ DSTEA}}}}\end{tabular}}} & \multicolumn{1}{l}{\vcell{STAR}}  & \multicolumn{1}{l}{\vcell{\begin{tabular}[b]{@{}l@{}}STAR\\\textbf{\textbf{\textbf{\textbf{+ DSTEA}}}}\end{tabular}}}  \\[-\rowheight]
\printcellbottom       & \multicolumn{1}{l}{\printcelltop}   & \multicolumn{1}{r|}{\printcellmiddle}                                                                                                                         & \multicolumn{1}{l}{\printcelltop}  & \multicolumn{1}{l|}{\printcellmiddle}                                                                                     & \multicolumn{1}{l}{\printcelltop} & \multicolumn{1}{l|}{\printcellmiddle}                                                                                   & \multicolumn{1}{l}{\printcelltop} & \multicolumn{1}{l}{\printcellmiddle}                                                                                   \\ 
\hline
attraction-area        & 91.37                               & \textbf{92.09}                                                                                                                                                & 86.79                              & \textbf{86.84}                                                                                                            & 88.33                             & 87.37                                                                                                                   & 90.41                             & 89.37                                                                                                                  \\
attraction-name        & 83.86                               & \textbf{85.34}                                                                                                                                                & 76.99                              & \textbf{79.88}                                                                                                            & 70.36                             & \textbf{71.66}                                                                                                          & 69.55                             & \textbf{70.79}                                                                                                         \\
attraction-type        & 89.33                               & \textbf{91.02}                                                                                                                                                & 85.47                              & \textbf{87.01}                                                                                                            & 85.39                             & \textbf{86.00}                                                                                                          & 85.78                             & 85.17                                                                                                                  \\
hotel-area             & 87.67                               & 85.45                                                                                                                                                         & 74.42                              & \textbf{77.15}                                                                                                            & 74.77                             & \textbf{76.69}                                                                                                          & 75.95                             & \textbf{78.14}                                                                                                         \\
hotel-book day         & 96.37                               & \textbf{97.05}                                                                                                                                                         & 94.21                              & 92.91                                                                                                                     & 95.87                             & 95.60                                                                                                                   & 95.19                             & \textbf{95.47}                                                                                                         \\
hotel-book people      & 98.15                               & 97.96                                                                                                                                                         & 88.88                              & \textbf{91.34}                                                                                                            & 95.03                             & \textbf{96.21}                                                                                                          & 95.17                             & \textbf{96.39}                                                                                                         \\
hotel-book stay        & 96.75                               & \textbf{97.03}                                                                                                                                                & 94.62                              & 94.25                                                                                                                     & 95.95                             & 94.50                                                                                                                   & 96.31                             & \textbf{96.69}                                                                                                         \\
hotel-internet         & 83.55                               & \textbf{86.93}                                                                                                                                                & 79.51                              & 77.67                                                                                                                     & 82.21                             & 80.89                                                                                                                   & 78.45                             & \textbf{80.30}                                                                                                         \\
hotel-name             & 90.44                               & \textbf{91.51}                                                                                                                                                & 79.38                              & \textbf{83.39}                                                                                                            & 77.62                             & \textbf{78.16}                                                                                                          & 78.59                             & \textbf{79.25}                                                                                                         \\
hotel-parking          & 85.72                               & 85.44                                                                                                                                                         & 77.03                              & \textbf{77.72}                                                                                                            & 79.19                             & \textbf{80.38}                                                                                                          & 82.67                             & 80.87                                                                                                                  \\
hotel-pricerange       & 89.20                               & 87.87                                                                                                                                                         & 80.02                              & \textbf{84.95}                                                                                                            & 84.95                             & \textbf{87.95}                                                                                                          & 83.72                             & \textbf{86.03}                                                                                                         \\
hotel-stars            & 89.89                               & 89.22                                                                                                                                                         & 87.23                              & 85.96                                                                                                                     & 87.20                             & \textbf{87.95}                                                                                                          & 88.42                             & 86.87                                                                                                                  \\
hotel-type             & 74.47                               & 74.01                                                                                                                                                         & 65.42                              & \textbf{68.21}                                                                                                            & 71.63                             & 71.34                                                                                                                   & 69.90                             & 67.44                                                                                                                  \\
restaurant-area        & 93.35                               & \textbf{94.63}                                                                                                                                                & 88.71                              & 88.03                                                                                                                     & 90.07                             & \textbf{91.35}                                                                                                          & 90.42                             & \textbf{91.55}                                                                                                         \\
restaurant-book day    & 98.39                               & \textbf{98.70}                                                                                                                                                & 95.72                              & 94.72                                                                                                                     & 97.05                             & \textbf{97.74}                                                                                                          & 96.97                             & \textbf{97.89}                                                                                                         \\
restaurant-book people & 97.74                               & \textbf{98.64}                                                                                                                                                & 92.43                              & \textbf{93.96}                                                                                                            & 95.76                             & \textbf{96.26}                                                                                                          & 96.06                             & \textbf{97.26}                                                                                                         \\
restaurant-book time   & 96.26                               & \textbf{96.72}                                                                                                                                                & 95.62                              & 94.86                                                                                                                     & 95.84                             & 94.62                                                                                                                   & 96.22                             & 95.92                                                                                                                  \\
restaurant-food        & 93.21                               & \textbf{95.57}                                                                                                                                                & 91.71                              & \textbf{92.81}                                                                                                            & 92.07                             & \textbf{92.30}                                                                                                          & 91.28                             & \textbf{91.79}                                                                                                         \\
restaurant-name        & 86.56                               & \textbf{86.87}                                                                                                                                                & 84.69                              & 82.95                                                                                                                     & 74.63                             & 74.49                                                                                                                   & 75.55                             & 71.43                                                                                                                  \\
restaurant-pricerange  & 92.82                               & \textbf{94.15}                                                                                                                                                & 89.25                              & \textbf{91.05}                                                                                                            & 91.34                             & \textbf{92.41}                                                                                                          & 90.04                             & \textbf{91.21}                                                                                                         \\
taxi-arriveby          & 85.40                               & \textbf{92.70}                                                                                                                                                & 83.14                              & 79.28                                                                                                                     & 74.82                             & \textbf{78.38}                                                                                                          & 69.92                             & \textbf{79.68}                                                                                                         \\
taxi-departure         & 78.08                               & \textbf{81.33}                                                                                                                                                & 57.95                              & 52.72                                                                                                                     & 79.15                             & \textbf{79.77}                                                                                                          & 81.74                             & \textbf{84.32}                                                                                                         \\
taxi-destination       & 81.26                               & \textbf{85.68}                                                                                                                                                & 66.17                              & 57.87                                                                                                                     & 82.21                             & 81.51                                                                                                                   & 83.58                             & \textbf{84.82}                                                                                                         \\
taxi-leaveat           & 87.80                               & 86.28                                                                                                                                                         & 82.05                              & \textbf{82.28}                                                                                                            & 80.97                             & \textbf{81.28}                                                                                                          & 76.01                             & \textbf{83.23}                                                                                                         \\
train-arriveby         & 87.35                               & \textbf{87.42}                                                                                                                                                & 85.53                              & \textbf{85.88}                                                                                                            & 86.33                             & 85.66                                                                                                                   & 85.42                             & \textbf{87.13}                                                                                                         \\
train-bookpeople       & 89.04                               & \textbf{89.22}                                                                                                                                                & 83.44                              & \textbf{85.25}                                                                                                            & 90.65                             & \textbf{90.98}                                                                                                          & 85.33                             & \textbf{85.60}                                                                                                         \\
train-day              & 97.84                               & \textbf{99.13}                                                                                                                                                & 97.92                              & 97.40                                                                                                                     & 98.45                             & \textbf{98.68}                                                                                                          & 98.75                             & \textbf{98.79}                                                                                                         \\
train-departure        & 96.83                               & \textbf{97.45}                                                                                                                                                & 94.45                              & 94.42                                                                                                                     & 95.44                             & 94.86                                                                                                                   & 96.16                             & \textbf{96.47}                                                                                                         \\
train-destination      & 97.60                               & \textbf{98.94}                                                                                                                                                & 96.53                              & \textbf{96.56}                                                                                                            & 97.14                             & 96.76                                                                                                                   & 96.10                             & \textbf{97.11}                                                                                                         \\
train-leaveat          & 68.89                               & 67.44                                                                                                                                                         & 67.35                              & \textbf{67.84}                                                                                                            & 68.85                             & 67.78                                                                                                                   & 67.22                             & \textbf{68.17}                                                                                                         \\
\Xhline{3\arrayrulewidth}
\end{tabular}}
\caption{Slot-specific performance on the test sets of MultiWOZ 2.1. The score of joint goal accuracy for each slot is computed. If the \textbf{+ DSTEA} performance is improved compared to the baseline, it is shown in bold. }
\label{tab:domain_2.1}
\end{table*}

\begin{table*}
\centering
\resizebox{0.6\textwidth}{!}{%
\setlength{\tabcolsep}{10pt}
\renewcommand{\arraystretch}{1.5}
\begin{tabular}{l|cc|cc} 
\Xhline{3\arrayrulewidth}
\multicolumn{5}{c}{MultiWOZ 2.2}                                                                                                                                                                                                                                                                                                                                                              \\ 
\hline
\vcell{Slot}           & \multicolumn{1}{l}{\vcell{SOM-DST}} & \multicolumn{1}{r|}{\vcell{\begin{tabular}[b]{@{}r@{}}SOM-DST\textbf{\textbf{\textbf{\textbf{}}}}\\\textbf{\textbf{\textbf{\textbf{+ DSTEA}}}}\end{tabular}}} & \multicolumn{1}{l}{\vcell{Trippy}} & \multicolumn{1}{l}{\vcell{\begin{tabular}[b]{@{}l@{}}Trippy \\\textbf{\textbf{\textbf{\textbf{+ DSTEA}}}}\end{tabular}}}  \\[-\rowheight]
\printcellbottom       & \multicolumn{1}{l}{\printcelltop}   & \multicolumn{1}{r|}{\printcellmiddle}                                                                                                                         & \multicolumn{1}{l}{\printcelltop}  & \multicolumn{1}{l}{\printcellmiddle}                                                                                      \\ 
\hline
attraction-area        & 90.89                               & \textbf{92.58}                                                                                                                                                & 85.36                              & \textbf{88.67}                                                                                                            \\
attraction-name        & 84.15                               & 83.64                                                                                                                                                         & 81.53                              & 79.20                                                                                                                     \\
attraction-type        & 89.33                               & 89.15                                                                                                                                                         & 86.35                              & 85.70                                                                                                                     \\
hotel-area             & 87.59                               & \textbf{87.91}                                                                                                                                                & 72.57                              & \textbf{76.88}                                                                                                            \\
hotel-book day         & 96.85                               & 96.56                                                                                                                                                         & 94.61                              & \textbf{94.99}                                                                                                            \\
hotel-book people      & 97.18                               & \textbf{97.28}                                                                                                                                                & 94.56                              & 93.47                                                                                                                     \\
hotel-book stay        & 97.61                               & 97.13                                                                                                                                                         & 93.54                              & \textbf{94.73}                                                                                                            \\
hotel-internet         & 82.75                               & \textbf{88.53}                                                                                                                                                & 82.20                              & 80.23                                                                                                                     \\
hotel-name             & 90.16                               & \textbf{91.09}                                                                                                                                                & 82.75                              & 81.84                                                                                                                     \\
hotel-parking          & 87.69                               & \textbf{88.26}                                                                                                                                                & 81.50                              & 80.31                                                                                                                     \\
hotel-pricerange       & 91.80                               & 90.89                                                                                                                                                         & 82.23                              & \textbf{84.06}                                                                                                            \\
hotel-stars            & 89.73                               & \textbf{91.33}                                                                                                                                                & 87.05                              & \textbf{87.61}                                                                                                            \\
hotel-type             & 76.95                               & \textbf{83.14}                                                                                                                                                & 66.27                              & \textbf{67.56}                                                                                                            \\
restaurant-area        & 94.19                               & 93.90                                                                                                                                                         & 88.68                              & 87.88                                                                                                                     \\
restaurant-book day    & 98.70                               & 98.62                                                                                                                                                         & 96.16                              & \textbf{96.37}                                                                                                            \\
restaurant-book people & 97.52                               & \textbf{99.47}                                                                                                                                                & 94.52                              & \textbf{96.19}                                                                                                            \\
restaurant-book time   & 95.88                               & 95.72                                                                                                                                                         & 95.69                              & \textbf{95.77}                                                                                                            \\
restaurant-food        & 95.62                               & 94.85                                                                                                                                                & 92.32                              & 92.28                                                                                                                     \\
restaurant-name        & 83.98                               & \textbf{84.96}                                                                                                                                                & 81.68                              & 80.83                                                                                                                     \\
restaurant-pricerange  & 93.18                               & \textbf{94.15}                                                                                                                                                & 89.91                              & \textbf{90.75}                                                                                                            \\
taxi-arriveby          & 88.41                               & 83.26                                                                                                                                                         & 78.26                              & \textbf{79.52}                                                                                                            \\
taxi-departure         & 74.60                               & \textbf{75.82}                                                                                                                                                & 53.88                              & \textbf{59.25}                                                                                                            \\
taxi-destination       & 80.57                               & \textbf{81.60}                                                                                                                                                & 64.83                              & 64.68                                                                                                                     \\
taxi-leaveat           & 88.10                               & 86.89                                                                                                                                                         & 81.73                              & \textbf{82.60}                                                                                                            \\
train-arriveby         & 86.91                               & 86.85                                                                                                                                                         & 85.62                              & \textbf{85.77}                                                                                                            \\
train-bookpeople       & 88.69                               & \textbf{90.18}                                                                                                                                                & 81.98                              & \textbf{85.81}                                                                                                            \\
train-day              & 99.16                               & 99.09                                                                                                                                                         & 97.51                              & \textbf{98.11}                                                                                                            \\
train-departure        & 97.10                               & 96.87                                                                                                                                                         & 95.29                              & 94.60                                                                                                                     \\
train-destination      & 97.61                               & \textbf{98.36}                                                                                                                                                & 96.11                              & \textbf{96.46}                                                                                                            \\
train-leaveat          & 68.16                               & \textbf{68.56}                                                                                                                                                & 66.05                              & \textbf{66.89}                                                                                                            \\
\Xhline{3\arrayrulewidth}
\end{tabular}}
\caption{Slot-specific performance on the test sets of MultiWOZ 2.2. The score of joint goal accuracy for each slot is computed. If the \textbf{+ DSTEA} performance is improved compared to the baseline, it is shown in bold.  }
\label{tab:domain_2.2}
\end{table*}

\end{document}